\documentclass[letterpaper]{article} 
\usepackage[preprint]{aaai2027}  
\usepackage[hyphens]{url}  
\usepackage{graphicx} 
\usepackage{natbib}  
\usepackage{caption} 
\usepackage{algorithm}
\usepackage{algorithmic}

\usepackage{newfloat}
\usepackage{listings}
\DeclareCaptionStyle{ruled}{labelfont=normalfont,labelsep=colon,strut=off} 
\floatstyle{ruled}
\newfloat{listing}{tb}{lst}{}
\floatname{listing}{Listing}

\usepackage{booktabs}

\title{Every Little Helps: Building Knowledge Graph Foundation Model with Fine-grained Transferable Multi-modal Tokens}
\author{
    Yichi Zhang\textsuperscript{\rm 1},
    Zhuo Chen\textsuperscript{\rm 1},
    Lingbing Guo\textsuperscript{\rm 2},
    Wen Zhang\textsuperscript{\rm 1},
    Huajun Chen\textsuperscript{\rm 1}
}
\affiliations{
    \textsuperscript{\rm 1}Zhejiang University, \textsuperscript{\rm 2}Nanjing University\\
}

\usepackage{amsmath}
\usepackage{mathtools}
\usepackage{amsthm}
\usepackage{multirow}
\usepackage{bm}
\newcommand{\model}{TOFU}

\begin{document}

\maketitle

\begin{abstract}
\textbf{Multi-modal knowledge graph reasoning} (MMKGR) aims to predict the missing links by exploiting both graph structure information and multi-modal entity contents. Most existing works are designed for a transductive setting, which learns dataset-specific embeddings and struggles to generalize to new KGs. Recent knowledge graph foundation models (KGFMs) improve cross-KG transfer, but they mainly exploit structural patterns and ignore rich multi-modal signals. We address these gaps by proposing a token-based foundation model ({\model}) for MMKGR, which exhibits strong generalization across different MMKGs. {\model} discretizes structural, visual, and textual information into modality-specific tokens. {\model} then employs a hierarchical fusion architecture with mixture-of-messages mechanisms, aiming to process these tokens and obtain transferable features for MMKGR. Experimental results on 17 transductive, inductive, and fully-inductive MMKGs show that {\model} generally outperforms strong KGFM and MMKGR baselines, delivering strong performance on unseen MMKGs.
\end{abstract}

\section{Introduction}
Multi-modal knowledge graph reasoning (MMKGR) \cite{MMKG-Survey} aims to discover new relational links among multi-modal entities. Unlike traditional KGR, MMKGR must model complex relational patterns \cite{bordes_translating_2013-TransE} while effectively integrating heterogeneous modality information (e.g., images and texts), which makes model design challenging and has attracted increasing research attention.
\begin{figure}[h]
  \centering
  \includegraphics[width=0.85\linewidth]{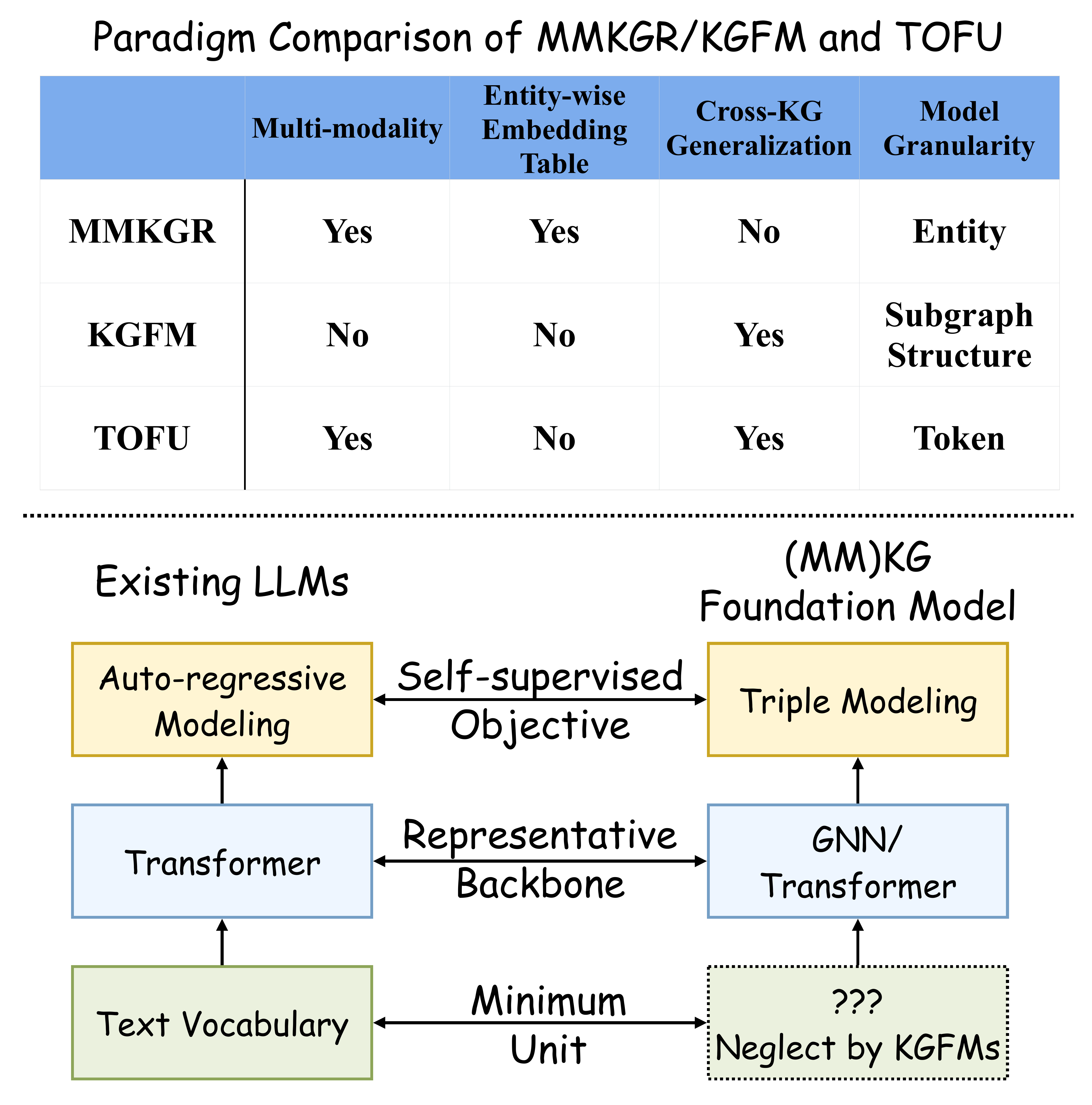}
  \caption{A comparison between existing MMKGR/KGFM and {\model}, along with insights from large language models.} 
  \label{fig::introduction}
  \vspace{-16pt}
\end{figure}
\par Most MMKGR methods \cite{LMBKGC,MyGO,APKGC} are developed under a transductive setting: they train and evaluate on the same MMKG with fixed entity and relation sets, relying on dataset-specific entity and relation embeddings. These models perform well on the training KG but generalize poorly to unseen entities, unseen relations, or new KGs, which limits their reusability. Recently, more works are considering the KG foundation model (KGFM) \cite{ULTRA, KGICL}, which is inspired by the paradigm of large foundation models \cite{llama} that have strong generalization capability. However, these KGFMs mainly extract transferable structural patterns, largely ignoring richer multi-modal signals, even if existing research \cite{xie_image-embodied_2017-IKRL} indicates that multi-modal contents significantly aid reasoning outcomes. This naturally raises a question: \textbf{\textit{Can we build KGFM by considering multi-source information from both structure and multi-modal contents?}}

\par To achieve this goal, two key challenges arise: (C1) How to define transferable features across different modalities? (C2) How to design effective architectures that fully exploit these transferable features? As illustrated in Figure~\ref{fig::introduction}, existing MMKGR studies rarely consider cross-KG transferability of multi-modal information, while various KGFM approaches focus on exploring structural features. Current MMKGR work primarily relies on shallow embeddings, whereas KGFM mainly leverages graph neural networks (GNNs) \cite{NBFNet} to learn structural features, lacking exploration on transferable multi-modal fusion.
\par We draw inspiration from the successful foundation models like LLMs. LLMs convert raw text into a fixed vocabulary, using subword-level tokens to convert text into tokens as the minimum units and basics of the foundation model, and employ representative neural network backbones such as transformers \cite{transformer} and MoE \cite{Deepseek-MoE}. Finally, a self-supervised training objective is defined for large-scale pre-training. For language, the specific objective is the auto-regressive profile for texts. The key idea is that a stable tokenization scheme plus a unified backbone yields a reusable model that can be applied across diverse tasks and domains. The vocabulary serves as a self-contained module providing unified text processing capabilities. We posit that a similar token-based design can also benefit MMKGR.

\par Therefore, we propose a \underline{\textbf{TO}}ken-based MMKG \underline{\textbf{F}}o\underline{\textbf{U}}ndation model ({\model}). {\model} builds a KGFM with fine-grained transferable multi-modal tokens by treating all modalities, including structural, visual, and textual information, as discrete token sequences. For visual and textual information, {\model} borrows tokenizers and pre-trained codebooks from the vision/text foundation models \cite{BERT} to process the modality information into token-level embedding sequences. For the structural modality, {\model} encodes entities into discrete tokens based on their relative positions in sampled context subgraphs. {\model} further employs a hierarchical fusion architecture that consists of a structural encoder and a multi-modal encoder to obtain the transferable modality features, which are fused with a relation-aware adaptive gate. Through this module, we obtain transferable entity and relation representations. {\model} further proposes a global aggregation module that propagates compositional multi-modal messages via a mixture-of-messages mechanism to perform multi-modal triple modeling on context subgraphs. Importantly, {\model} removes entity-/relation-specific embedding tables and instead learns to utilize transferable multi-modal features, enabling a single model to be pre-trained and fine-tuned across different MMKGs with a unified objective, and to support both zero-shot and supervised fine-tuning transfer. To verify the effectiveness of {\model}, we conduct comprehensive experiments on 17 MMKGs with transductive, inductive, and fully-inductive settings. Through experimentation, we demonstrate that {\model} exhibits robust zero-shot capabilities after pretraining, which further improve upon fine-tuning and surpass baseline performance. We also provide in-depth analyses and visualizations to explain the behavior of key components and to justify our designs. Our contribution can be summarized as:
\par \textbf{(1). New Perspective for MMKGR.} Motivated by current LLMs, we first propose a token-based paradigm to build a foundation model for MMKGR. This paradigm processes different modalities into discrete token representations through specific methods and models them unifiedly.
\par \textbf{(2). Foundation Architecture.} We introduce {\model}, a token-based foundation architecture for MMKGR that combines a hierarchical fusion module and a global propagation module to capture transferable multi-modal features from both local and global views. {\model} can complete pre-training and fine-tuning with a unified objective, and induce new entities and relations across different MMKGs.
\par \textbf{(3). Extensive Experiments.} We conduct comprehensive experiments on 17 MMKGs with transductive, inductive, and fully-inductive settings for performance evaluation. Further explorations are made to show the rationality and interpretability of {\model}.

\section{Related Works}
\begin{figure*}
  \centering
  \includegraphics[width=\linewidth]{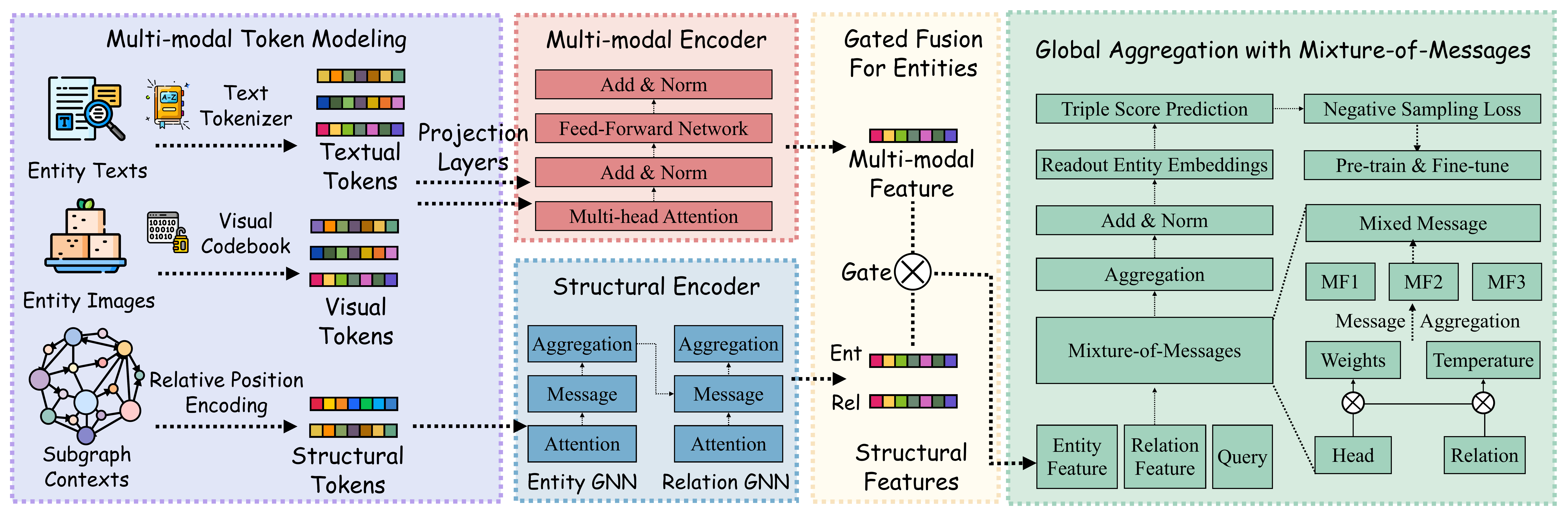}
  \vspace{-16pt}
  \caption{An overview of our {\model} framework. {\model} first models each modality (structure/visual/textual) into discrete tokens and employs a hierarchical fusion architecture to obtain the transferable entity and relation features, which consists of a structural encoder, a multi-modal encoder, and a fusion gate. Finally, {\model} applies global aggregation with a mixture-of-message mechanism to obtain multi-source information from the MMKG to make the MMKGR prediction.} 
  \label{fig::introduction}
  \vspace{-8pt}
\end{figure*}

\noindent \textbf{Foundation Model for KG.}
Traditional KG reasoning models learn separate embeddings for entities and relations in the static KG, which can not be generalized to new entities and relations. Knowledge graph foundation model (KGFM) \cite{KGICL} aims to build a foundation model for KG reasoning, which possesses cross-KG transfer capabilities, enabling generalization to new knowledge graphs with unseen entities and relations during training. ULTRA \cite{ULTRA} proposes relative relation representations as transferable features to build a foundation model. MOTIF \cite{MOTIF} further extends ULTRA to a relational hyper-graph to learn the relation patterns for generalization. KG-ICL \cite{KGICL} modeling the generalizable features with in-context graph prompts. These methods focus on structural information as transferable features, which neglect the importance of multi-modal information in the KG. Some MMKG models like IndMKG \cite{IndMKG} have explored transfer learning of multi-modal information, but have not pursued further research along the KGFM pathway. 

\section{Preliminaries}
\textbf{MMKG Reasoning (MMKGR).} MMKGs can be represented as $\mathcal{KG}=(\mathcal{E}, \mathcal{R}, \mathcal{T}, \mathcal{M})$ where $\mathcal{E}, \mathcal{R}$ are the entity and relation set. $\mathcal{T}=\{(h, r, t)|h, t\in\mathcal{E}, r\in\mathcal{R}\}$ consists of structured triple represent factual knowledge. $\mathcal{M}=(\mathcal{M}_{vis}, \mathcal{M}_{txt})$ represents the multi-modality content set of the KG. In this paper, we mainly focus on two common modalities: visual (vis) and textual (txt). Given a query $(h, r, ?)$, MMKGR aims to predict the missing entity in the query by taking into account both the structural information in $\mathcal{T}$ and the multimodal content in $\mathcal{M}$. If the head entity $h$ and relation $r$ appearing in a query have already been seen during training, we refer to it as \textbf{\textit{transductive}} MMKGR; otherwise, it is referred to as \textbf{\textit{inductive}} MMKGR.

\noindent\textbf{KG Foundation Model.} KGFM aims to achieve transferable KGR by modeling the transferable features in the KGs. A KGFM can be trained on several KGs $\mathcal{D}_{train}=\{\mathcal{KG}_{train}^{(i)}\}_{i=1}^{m}$ and evalute on other KGs $\mathcal{D}_{test}=\{\mathcal{KG}^{(i)}_{test}\}_{i=1}^{n}$, which correspond to the original KG's splits, so the training and testing data may share the same entity/relation sets (transductive) or may not (inductive). This places higher demands on KGFM's generalization capabilities. In this paper, all KGs considered are MMKGs with additional multi-modal information including texts and images.

\section{Methodology}
In this section, we present the details of the {\model} framework. We first describe how {\model} captures fine-grained transferable features from MMKGs and then encodes them into local- and global-aware multi-modal knowledge representations for MMKGR.
\subsection{Transferable Multi-modal Tokens}
Compared with traditional KGR methods, the key advantage of KGFMs lies in their ability to capture transferable features and patterns that can be generalized across different KGs. In {\model}, we propose a token-based transferable modeling approach, where different modalities are uniformly represented as discrete tokens.
\subsubsection{Visual and Textual Tokens}
The multi-modal contents (visual and textual information) of entities can be easily processed into discrete tokens. Given an entity $e$, its multi-modal contents $\mathcal{M}_{vis}(e), \mathcal{M}_{txt}(e)$ can be process into a token sequence as:
\begin{equation}
    \mathcal{X}_m(e)=\mathcal{Q}_m(\mathcal{M}_m(e))=[x_{m, 1}, x_{m, 1}, \cdots, x_{m, n_m}]
\end{equation}
where $\mathcal{Q}_{m}$ is the processor for modality $m\in\{txt, vis\}$, transfering entity modality content into $n_m$ discrete tokens. $\mathcal{Q}_{m}$ is a text tokenizer or a pre-trained VQ-VAE network \cite{VQ-VAE} for visual modality. $x_{m, i}$ are the corresponding vectors of the semantic ids in the codebook of $\mathcal{Q}_m$. Compared to existing MMKGR methods, this token-based approach offers more fine-grained multimodal features for subsequent modeling. Since these features are derived from a fixed pre-trained codebook, they exhibit strong generalization ability across different entities.

\subsubsection{Structural Token Modeling}
In contrast to visual and textual contents, the structural modality cannot be directly extracted from an entity in isolation. Traditional KGR methods usually define entity-wise learnable embeddings to model structural information, which is simple but lacks transferability to unseen entities. Following existing KGFMs, {\model} employs relative position tokens to capture entity structures within a subgraph. Given a triple $(h, r, t)$, a subgraph $\mathcal{KG}'=(\mathcal{E}', \mathcal{R}', \mathcal{T}')$ is randomly sampled from the original $\mathcal{KG}$ with the $k$-hop neighbors of $(h, r, t)$. For every $e\in\mathcal{E}'$, we compute its shortest-path distances to the head and tail entities, forming a positional tuple $[d(e,h), d(e,t)]$, where $d(\cdot)$ is the distance function. This tuple, which represents the entity's relative structural role, is then mapped to a corresponding learnable embedding. These embeddings effectively form a discrete codebook of structural tokens, analogous to the visual and textual tokens.

\subsection{Hierarchical Local Fusion}
Next, we propose a hierarchical architecture that enables multi-level, multi-modal fusion and local feature aggregation for entities. A key distinction is that while our visual and textual tokens are derived from pre-trained models, the structural tokens are initialized without pre-trained weights. Therefore, we first develop a structural encoder (SE) to learn contextualized structural representations for entities and relations within their local subgraph contexts.
\subsubsection{Structural Encoder}
\label{main::structural_encoder}
The structural encoder is tasked with capturing transferable structural patterns from the subgraph context. As previously described, structural information is represented by a learnable codebook of position embeddings. Therefore, we employ an $ L_1$-layer graph neural network (GNN) as the subgraph structural encoder (SE). Given a query $q=(h, r, ?)$ and its subgraph context $\mathcal{KG}'$, we assign the entity and relation hidden representations in the $i$-th layer of the SE as $\mathcal{H}_{ent}^{(i)},\mathcal{H}_{rel}^{(i)}$. Here, $\mathcal{H}_{ent}^{(0)}$ is the structural token embeddings and $\mathcal{H}_{rel}^{(0)}$ is initialized by zero vectors. A learnable query vector shared by all relations is added to the query relation $r$. In SE, the entity and relation representations are iteratively updated as:
\begin{equation}
    \mathcal{H}_{ent}^{(i+1)}=\mathbf{AGG}_{n\in\mathcal{N}(e)}(\mathbf{MSG}(\mathcal{H}_{ent}^{(i)}, \mathcal{H}_{rel}^{(i)}, n, q))
\end{equation}

\begin{equation}
    \mathcal{H}_{rel}^{(i+1)}=\mathbf{AGG}_{n\in\mathcal{N}(r)}(\mathbf{MSG}(\mathcal{H}_{ent}^{(i)}, \mathcal{H}_{rel}^{(i)}, n, q))
\end{equation}
Here, $\mathcal{N}(e), \mathcal{N}(r)$ represent the neighbors of the entities and relations in $\mathcal{KG}'$. $\mathbf{AGG}, \mathbf{MSG}$ are the aggregation and message functions in the GNN, respectively. We concatenate the query $q$ and hidden representations and project it with one MLP as the message:
\begin{equation}
    \mathbf{MSG}(\bm{h}^{(i)}, \bm{r}^{(i)}, \bm{t}^{(i)}, q)=\alpha_{r,q}\mathcal{P}_{ent}^{(i)}\left([\bm{h}^{(i)}, \bm{r}^{(i)},q]\right)
\end{equation}
\begin{equation}
    \alpha_{ r,q}=\sigma\left(\mathcal{P}_{attn}^{(i)}\left([\bm{r}^{(i)},q]\right)\right)
\end{equation}
This is a micro-view of the $\mathbf{MSG}$ function on a single edge (triple) in the subgraph. $\alpha_{r,q}$ is the adaptive attention weights and $\mathcal{P}_{ent}^{(i)},\mathcal{P}_{attn}^{(i)}$ are projection layers. The messages would be further aggregated with a max-pooling layer with ReLU activation. Finally, we can obtain the structural entity and relation representations in the subgraph contexts from SE, which can be denoted as $\mathcal{H}_{ent}^{(L_1)}$, $\mathcal{H}_{rel}^{(L_1)}$. These features consist of structural representations based on relative positions and would be further used in the subsequent modules.

\subsubsection{Multi-modal Encoder and Gated Fusion}
We further employ a multi-modal encoder (ME) to encode the multi-modal contents for entities. We employ a transformer encoder \cite{transformer} with multi-head attention (MHA) and feed-forward network (FFN) modules to capture the multi-modal feature of entities as:
\begin{equation}
    f_{mm}(e)=\mathbf{Transformer}([\mathrm{ENT}], \mathcal{X}_{mm}(e))
\end{equation}
\begin{equation}
    \mathcal{X}_{mm}(e)=[\mathcal{P}_{txt}(\mathcal{X}_{txt}(e)), \mathcal{P}_{vis}(\mathcal{X}_{vis}(e)))]
\end{equation}
where $[\mathrm{ENT]}$ is a learnable readout token shared by all entities. We first project the visual and textual tokens with projection layers $\mathcal{P}_{\text{txt}}$ and $\mathcal{P}_{\text{vis}}$ to align their dimensions with the transformer input, and then take the output representation of $[\mathrm{ENT}]$ as the multi-modal representation of entity $e$. In summary, we can obtain two feature vectors $f_{str}(e), f_{mm}(e)$ for entity $e$, which represent its structural features in the subgraph contexts and multi-modal features from its raw contents. These two features are fused into a unified representation for subsequent prediction. We employ a gated-fusion (GF) module to achieve adaptive fusion as:
\begin{equation}
    f_{all}(e)=g_{str}\odot f_{str}(e) + g_{mm}\odot f_{mm}(e)
\end{equation}
\begin{equation}
    g_{str},g_{mm}=\sigma(\mathcal{P}_{all}([f_{str}(e), f_{mm}(e)]))
\end{equation}
Here we employ another projection layer $\mathcal{P}_{all}$ and sigmoid function $\sigma$ to project the concatenated structural and multi-modal features into adaptive gate weights.

\subsection{Global Propagation with Mixture-of-Messages}
In the above process, we achieve transferable entity and relation modeling with fine-grained tokens from three modalities. We further design the global propagation (GP) module to propagate the multi-modal representations to candidate entities for final prediction. GP is a $L_2$-layers GNN that also updates the entity representations iteratively as:
\begin{equation}
\widehat{\mathcal{H}}_{ent}^{(j+1)}=\mathbf{AGG}_{n\in\mathcal{N}(r)}(\mathbf{MSG}(\widehat{\mathcal{H}}_{ent}^{(j)}, \widehat{\mathcal{H}}_{rel}^{(j)}, n, q))
\end{equation}
For a given query $q=(h,r,?)$, the new entity representation matrix $\widehat{\mathcal{H}}_{ent}^{(0)}$ is initialized with $f_{all}$ for the query entity, and others are assigned as zero vectors. $\widehat{\mathcal{H}}_{rel}^{(0)}$ is initialized with the output of SE. Moreover, $\mathbf{MSG}$ and $\mathbf{AGG}$ are another group of message and aggregation functions. In {\model}, we propose a mixture-of-messages (MiM) module as $\mathbf{MSG}$, which is designed as:
\begin{equation}
    m_{h\rightarrow t}=\sum_{i=1}^{k}\beta_i m_{h\rightarrow t, i}\quad \beta_i=\frac{\exp(\gamma_i/\tau_i)}{\sum_{j=1}^{k}\exp (\gamma_j/\tau_j)}
\end{equation}
\begin{equation}
    \gamma_i=\mathcal{P}_{1}([\widehat{\mathcal{H}}_{ent}(h),\widehat{\mathcal{H}}_{rel}(r)])\quad\tau_i=\mathcal{P}_{2}(\widehat{\mathcal{H}}_{rel}(r))
\end{equation}
Here, we employ a series of heterogeneous message functions $m_{h\rightarrow t, i}$ combined by weight $\beta_i$, which is defined by $\gamma_i$ and $\tau_i$. $\gamma_i$ is the gate logits estimated with a projection layer and $\tau_i$ is the relation-aware temperature. With such a design, the weights for $m_{h\rightarrow t, i}$ can be dynamically adjusted by the head and relation (query) contexts. This is because we found that different message functions excel at handling different relations. Therefore, we designed MiM to dynamically combine various message functions, enabling them to adapt to different relational information and enhance their generalization capabilities across diverse KGs. In practice, the different message functions can be denoted as:
\begin{equation}
    m_{h\rightarrow t, i}=\widehat{\mathcal{H}}_{ent}(h) \otimes_i\widehat{\mathcal{H}}_{rel}(r)
\end{equation}
Here, $\otimes_i$ is the operator for the $i$-th message function. We employ three classic KGR model TransE \cite{bordes_translating_2013-TransE}, Distmult \cite{yang_embedding_2015-DistMult}, and RotatE \cite{sun_rotate_2019-RotatE} as $\otimes_i$, which are addition, element-wise product, and complex rotation, respectively. After message passing, {\model} further aggregates the messages with a sum operation. A score layer $\mathcal{P}_{score}$ is defined to calculate the score of a candidate $t$ for the query $(h, r, ?)$ as:
\begin{equation}
    \mathcal{S}(h, r, t)=\mathcal{P}_{score}(\widehat{\mathcal{H}}_{ent}^{(L_2)}(t))
\end{equation}
Note that the query encoded with multi-modal information has been propagated to all entities. Therefore, the final hidden representations of the query message can be directly used for KGR. The training objective can be denoted as:
\begin{equation}
    \mathcal{L}_{kgr}=-\sum_{(h, r, t)\in\mathcal{T}}\log\frac{\exp (\mathcal{S}(h, r, t))}{\sum_{t'\in\mathcal{E}}\exp (\mathcal{S}(h, r, t'))}
\end{equation}
Following traditional KGR settings, all entities except the ground-truth one are treated as negative labels. Throughout the entire design, we do not define any entity-specific or relation-specific parameters. Instead, we build an MMKGR model with cross-KG generalization capabilities from the ground up using fine-grained multi-modal tokens. {\model} can be pre-trained on existing MMKGs to enable zero-shot inference on new MMKGs, or fine-tuned using the training data of new MMKGs. The training objective remains identical in both the pre-training and fine-tuning phases.
\begin{table}[]
\centering
\resizebox{0.48\textwidth}{!}{
\begin{tabular}{c|ccc|ccc|ccc}
\toprule
\multirow{2}{*}{\textbf{Dataset}} & \multicolumn{3}{c|}{\textbf{Train}} & \multicolumn{3}{c|}{\textbf{Valid}} & \multicolumn{3}{c}{\textbf{Test}} \\
 & $|\mathcal{E}|$ & $|\mathcal{R}|$ & $|\mathcal{T}|$ & $|\mathcal{E}|$ & $|\mathcal{R}|$ & $|\mathcal{T}|$ & $|\mathcal{E}|$ & $|\mathcal{R}|$ & $|\mathcal{T}|$ \\
\midrule
\multicolumn{10}{c}{\textbf{\textit{Transductive MMKGs}}} \\
\midrule
\textbf{DB15K} & 12842 & 279 & 79222 & 12842 & 279 & 9902 & 12842 & 279 & 9904 \\
\textbf{MKG-W} & 15000 & 169 & 34196 & 15000 & 169 & 4276 & 15000 & 169 & 4274 \\
\textbf{MKG-Y} & 15000 & 28 & 21310 & 15000 & 28 & 2665 & 15000 & 28 & 2663 \\
\textbf{YAGO15K} & 15283 & 32 & 86020 & 15283 & 32 & 12289 & 15283 & 32 & 24577 \\
\textbf{WN9} & 6555 & 9 & 11741 & 6555 & 9 & 1337 & 6555 & 9 & 1319 \\
\textbf{WN18RR++} & 41105 & 11 & 86835 & 41105 & 11 & 3034 & 41105 & 11 & 3134 \\
\textbf{FB15K-237-10\%} & 11512 & 237 & 27211 & 11512 & 237 & 15624 & 11512 & 237 & 18150 \\
\textbf{FB15K-237-20\%} & 13166 & 237 & 54423 & 13166 & 237 & 16963 & 13166 & 237 & 19776 \\
\textbf{FB15K-237-50\%} & 14149 & 237 & 136057 & 14149 & 237 & 17449 & 14149 & 237 & 20324 \\
\midrule
\multicolumn{10}{c}{\textbf{\textit{Inductive MMKGs}}} \\
\midrule
\textbf{FB15K-237(v1)} & 1594 & 180 & 4245 & 1594 & 180 & 489 & 1093 & 180 & 411 \\
\textbf{FB15K-237(v2)} & 3668 & 215 & 9799 & 3668 & 215 & 1166 & 2501 & 215 & 947 \\
\textbf{FB15K-237(v3)} & 3668 & 215 & 17986 & 3668 & 215 & 2194 & 2504 & 215 & 1731 \\
\textbf{FB15K-237(v4)} & 4707 & 219 & 27203 & 4707 & 219 & 3352 & 3051 & 219 & 2840 \\
\midrule
\multicolumn{10}{c}{\textbf{\textit{Fully-Inductive MMKGs}}} \\
\midrule
\textbf{FB-25} & 5190 & 163 & 91571 & 4097 & 216 & 5716 & 4097 & 216 & 5716 \\
\textbf{FB-50} & 5190 & 153 & 85375 & 4445 & 205 & 3879 & 4445 & 205 & 3879 \\
\textbf{FB-75} & 4659 & 134 & 62809 & 2792 & 186 & 3106 & 2792 & 186 & 3106 \\
\textbf{FB-100} & 4659 & 134 & 62809 & 2624 & 77 & 2329 & 2624 & 77 & 2329 \\
\bottomrule
\end{tabular}
}
\caption{Statistical informtion of the 17 MMKG datasets.}
\label{table::datasets_full}
\vspace{-8pt}
\end{table}
\section{Experiments}
\begin{table}[]
\centering
\resizebox{0.99\columnwidth}{!}{
\begin{tabular}{cc|cc|cc|cc|cc}
\toprule
\multicolumn{2}{c|}{\multirow{2}{*}{\textbf{Method}}} & \multicolumn{2}{c|}{\textbf{Transductive}} & \multicolumn{2}{c|}{\textbf{Inductive}} & \multicolumn{2}{c|}{\textbf{Fully-Inductive}} & \multicolumn{2}{c}{\textbf{Overall}} \\
\multicolumn{2}{c|}{} & \textbf{MRR} & \textbf{Hit@10} & \textbf{MRR} & \textbf{Hit@10} & \textbf{MRR} & \textbf{Hit@10} & \textbf{MRR} & \textbf{Hit@10} \\
\midrule
\multicolumn{2}{c|}{\textbf{Supervised SOTA}} & 42.47 & 54.70 & 47.73 & 63.58 & 16.55 & 29.63 & 37.61 & 50.89 \\
\midrule
\multirow{2}{*}{\textbf{ULTRA}} & zero-shot & 32.62 & 47.86 & 49.68 & 67.18 & 39.45 & 60.73 & 38.24 & 55.43 \\
 & fine-tune & 43.59 & 56.03 & 50.83 & 68.18 & 39.03 & 60.35 & 44.22 & 59.90 \\
 \midrule
\multirow{2}{*}{\textbf{KG-ICL}} & zero-shot & 41.70 & 52.09 & 53.33 & 70.53 & 41.05 & 63.55 & 44.28 & 59.13 \\
 & fine-tune & 41.83 & 52.68 & \underline{54.03} & \textbf{71.95} & \textbf{44.38} & \textbf{66.48} & 45.30 & 60.46 \\
 \midrule
\multirow{2}{*}{\textbf{MOTIF}} & zero-shot & 33.67 & 49.50 & 50.03 & 69.43 & 38.73 & 60.70 & 38.71 & 56.82 \\
 & fine-tune & 43.47 & \underline{56.01} & 52.85 & 70.63 & 37.93 & 58.98 & 44.37 & 60.15 \\
\midrule
\multirow{2}{*}{\textbf{\model}} & zero-shot & \underline{43.86} & 55.63 & 53.06 & 70.80 & 43.44 & \underline{66.15} & \underline{45.93} & \underline{61.67} \\
 & fine-tune & \textbf{46.12} & \textbf{57.92} & \textbf{54.49} & \underline{71.84} & \underline{43.22} & 65.68 & \textbf{47.41} & \textbf{63.02} \\
\toprule
\end{tabular}
}
\caption{The KGFM experiment results on 17 MMKGs.}
\label{table::main_kgfm}
\end{table}
\subsection{Experiment Settings}

\begin{table*}[]

\centering
\resizebox{\textwidth}{!}{
\begin{tabular}{c|cccc|cccc|cccc|cccc}
\toprule
\multirow{2}{*}{\textbf{Model}} & \multicolumn{4}{c|}{\textbf{DB15K}} & \multicolumn{4}{c|}{\textbf{MKG-W}} & \multicolumn{4}{c|}{\textbf{MKG-Y}} & \multicolumn{4}{c}{\textbf{Overall}} \\
 & \textbf{MRR} & \textbf{Hit@1} & \textbf{Hit@3} & \textbf{Hit@10} & \textbf{MRR} & \textbf{Hit@1} & \textbf{Hit@3} & \textbf{Hit@10} & \textbf{MRR} & \textbf{Hit@1} & \textbf{Hit@3} & \textbf{Hit@10} & \textbf{MRR} & \textbf{Hit@1} & \textbf{Hit@3} & \textbf{Hit@10} \\
\midrule
IKRL \cite{xie_image-embodied_2017-IKRL} & 26.82 & 14.09 & 34.93 & 49.09 & 32.36 & 26.11 & 34.75 & 44.07 & 33.22 & 30.37 & 34.28 & 38.26 & 30.80 & 23.52 & 34.65 & 43.81 \\
TBKGC \cite{sergieh_multimodal_2018-TBKGC} & 28.40 & 15.61 & 37.03 & 49.86 & 31.48 & 25.31 & 33.98 & 43.24 & 33.99 & 30.47 & 35.27 & 40.07 & 31.29 & 23.80 & 35.43 & 44.39 \\
TransAE \cite{wang_multimodal_2019-TransAE} & 28.09 & 21.25 & 31.17 & 41.17 & 30.00 & 21.23 & 34.91 & 44.72 & 28.10 & 25.31 & 29.10 & 33.03 & 28.73 & 22.60 & 31.73 & 39.64 \\
MMKRL \cite{DBLP:journals/apin/LuWJHL22-MMKRL} & 26.81 & 13.85 & 35.07 & 49.39 & 30.10 & 22.16 & 34.09 & 44.69 & 36.81 & 31.66 & 39.79 & 45.31 & 31.24 & 22.56 & 36.32 & 46.46 \\
RSME \cite{wang_is_2021-RSME} & 29.76 & 24.15 & 32.12 & 40.29 & 29.23 & 23.36 & 31.97 & 40.43 & 34.44 & 31.78 & 36.07 & 39.09 & 31.14 & 26.43 & 33.39 & 39.94 \\
VBKGC \cite{DBLP:journals/corr/abs-2209-07084-VBKGC} & 30.61 & 19.75 & 37.18 & 49.44 & 30.61 & 24.91 & 33.01 & 40.88 & 37.04 & 33.76 & 38.75 & 42.30 & 32.75 & 26.14 & 36.31 & 44.21 \\
OTKGE \cite{cao_otkge_2022-OTKGE} & 23.86 & 18.45 & 25.89 & 34.23 & 34.36 & 28.85 & 36.25 & 44.88 & 35.51 & 31.97 & 37.18 & 41.38 & 31.24 & 26.42 & 33.11 & 40.16 \\
MANS \cite{DBLP:conf/ijcnn/ZhangCZ23-MANS} & 28.82 & 16.87 & 36.58 & 49.26 & 30.88 & 24.89 & 33.63 & 41.78 & 29.03 & 25.25 & 31.35 & 34.49 & 29.58 & 22.34 & 33.85 & 41.84 \\
MMRNS \cite{MMRNS} & 32.68 & 23.01 & 37.86 & 51.01 & 35.03 & 28.59 & 37.49 & 47.47 & 35.93 & 30.53 & 39.07 & 45.47 & 34.55 & 27.38 & 38.14 & 47.98 \\
IMF \cite{li_imf_2023-IMF} & 32.25 & 24.20 & 36.00 & 48.19 & 34.50 & 28.77 & 36.62 & 45.44 & 35.79 & 32.95 & 37.14 & 40.63 & 34.18 & 28.64 & 36.59 & 44.75 \\
QEB \cite{DBLP:conf/mm/WangMCML023-TIVA} & 28.18 & 14.82 & 36.67 & 51.55 & 32.38 & 25.47 & 35.06 & 45.32 & 34.37 & 29.49 & 36.95 & 42.32 & 31.64 & 23.26 & 36.23 & 46.40 \\
VISTA \cite{lee_vista_2023-VISTA} & 30.42 & 22.49 & 33.56 & 45.94 & 32.91 & 26.12 & 35.38 & 45.61 & 30.45 & 24.87 & 32.39 & 41.53 & 31.26 & 24.49 & 33.78 & 44.36 \\
NATIVE \cite{NATIVE} & 34.30 & 25.08 & 39.48 & 51.35 & 36.84 & 29.94 & 40.06 & 49.39 & 39.21 & 35.03 & 41.21 & 46.25 & 36.78 & 30.02 & 40.25 & 49.00 \\
AdaMF \cite{AdaMF} & 32.51 & 21.31 & 39.67 & 51.68 & 34.27 & 27.21 & 37.86 & 47.21 & 38.06 & 33.49 & 40.44 & 45.48 & 34.95 & 27.34 & 39.32 & 48.12 \\
SNAG \cite{SNAG} & 36.30 & 27.40 & 41.10 & 53.00 & 37.30 & 30.20 & 40.50 & 50.30 & 39.10 & 34.70 & 41.08 & 46.70 & 37.57 & 30.77 & 40.89 & 50.00 \\
MyGO \cite{MyGO} & 37.72 & 30.08 & 41.26 & 52.21 & 36.10 & 29.78 & 38.54 & 47.75 & 38.44 & 35.01 & 39.84 & 44.19 & 37.42 & 31.62 & 39.88 & 48.05 \\
MoMoK \cite{MoMoK} & \underline{39.57} & \textbf{32.38} & \underline{43.45} & \underline{54.14} & 35.89 & 30.38 & 37.54 & 46.13 & 37.91 & 35.09 & 39.20 & 43.20 & 37.79 & 32.62 & 40.06 & 47.82 \\
K-ON \cite{K-ON} & 38.10 & 30.13 & 42.77 & 53.59 & 36.64 & 30.05 & 38.72 & 48.26 & 35.83 & 32.56 & 37.34 & 42.45 & 36.86 & 30.91 & 39.61 & 48.10 \\
MCKGC \cite{MCKGC} & \textbf{39.79} & \underline{31.92} & \textbf{43.80} & \textbf{54.66} & 36.88 & \underline{31.32} & 38.92 & 47.43 & 38.92 & \underline{35.49} & 40.57 & 45.21 & 38.53 & \underline{32.91} & 41.10 & 49.10 \\
APKGC \cite{APKGC} & 36.40 & 28.20 & 41.30 & 52.70 & \underline{37.40} & 30.60 & 40.40 & 50.10 & 38.41 & 34.52 & 40.27 & 45.02 & 37.40 & 31.11 & 40.66 & 49.27 \\
LMBKGC \cite{LMBKGC} & 37.23 & 27.78 & 42.75 & 54.71 & 38.46 & 31.20 & \underline{41.78} & \underline{51.46} & \underline{40.03} & 33.89 & \underline{43.11} & \textbf{50.81} & \underline{38.57} & 30.96 & \underline{42.55} & \textbf{52.33} \\
\midrule
\textbf{\model} & 39.01 & 31.80 & 42.34 & 53.14 & \textbf{40.40} & \textbf{34.39} & \textbf{42.98} & \textbf{51.66} & \textbf{43.72} & \textbf{40.50} & \textbf{45.25} & \underline{49.77} & \textbf{41.04} & \textbf{35.56} & \textbf{43.52} & \underline{51.52} \\
\bottomrule
\end{tabular}
}
\caption{Single-dataset transductive experiment results on three standard MMKGR benchmarks.}
\label{table::transductive_mmkgr}
\end{table*}
\par \textbf{Datasets.}
As shown in Table~\ref{table::datasets_full}, we collected and retained 17 MMKGs including: 9 transductive datasets, 4 inductive datasets, and 4 fully-inductive datasets.

\noindent \textbf{Baselines and Metrics.} For transductive experiments, we employ 21 recent SOTA MMKGC methods. For KGFM experiments, we compare our method with ULTRA, MOTIF, IndMKG, and KG-ICL with both pre-training and fine-tuning settings.  Following the classic KGC evaluation protocol, we report the Mean Reciprocal Rank (MRR) and Hit@K (K=1,3,10) results as evaluation metrics under filter setting \cite{bordes_translating_2013-TransE}. Implementation details of the experiments can be found in Appendix.
\noindent \textbf{Implementation Details.} For transductive experiments, we train and evaluate on the same MMKG. For other experiments, we pre-train {\model} on DB15K, MKG-W, MKG-Y, and FB15K-237(v1) and evaluate on 17 MMKGs as zero-shot performance. Besides, we fine-tune the pre-trained model on these MMKGs to obtain the fine-tune performance. This setting is consistent with the previous KGFM. All experiments are conducted on NVIDIA A100 GPUs.

\subsection{Main Experiments}
In the main experiments, we investigate the performance of {\model} as a KGFM from two complementary perspectives:

\noindent \textbf{(1). Pre-training and Fine-tuning on 17 MMKGs.} We first compare {\model} with several recent KGFM baselines under two settings: (i) zero-shot reasoning after pre-training, and (ii) supervised reasoning after task-specific fine-tuning. As summarized in Table~\ref{table::main_kgfm}, we report results for both settings across three standard scenarios: transductive, inductive, and fully-inductive. More fine-grained, dataset-wise results are provided in Table~\ref{table::kgfm_full}.

\par From these results, we observe that {\model} (i) achieves consistently strong performance under the transductive setting, showing its ability to capture rich structural patterns within a fixed graph; (ii) generalizes robustly in inductive and fully-inductive settings, attaining performance comparable to KG-ICL and approaching supervised SOTA results on several datasets; (iii) already surpasses several fine-tuned baselines in the zero-shot setting, especially on MMKGs with informative visual and textual modalities, demonstrating effective exploitation of multi-modal cues for cross-KG transfer; and (iv) maintains a balanced performance across 17 MMKGs and three evaluation settings, indicating its advantage as a general-purpose KG foundation model rather than a model tailored to a specific scenario.
\begin{figure}
  \centering
  \includegraphics[width=0.95\linewidth]{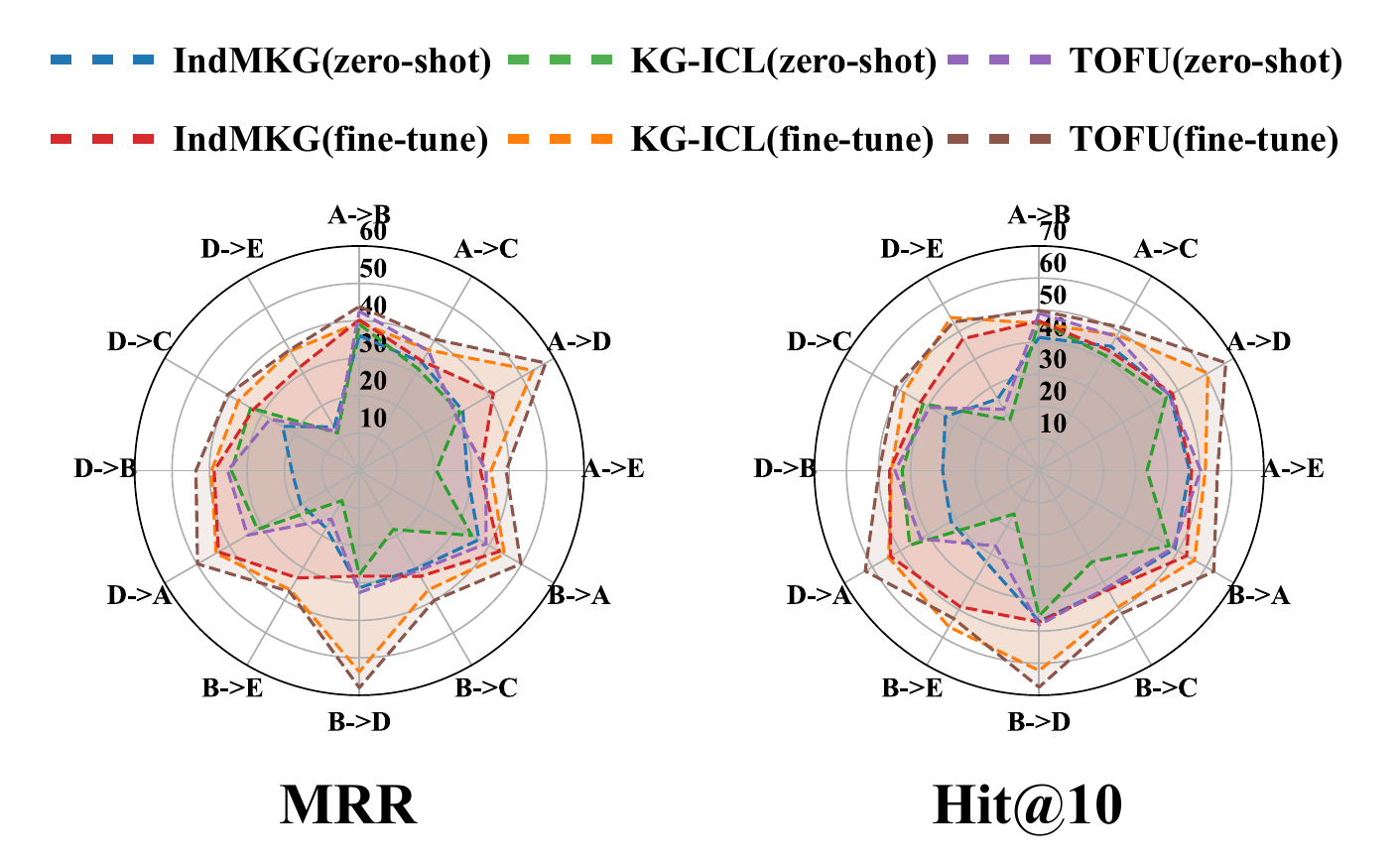}
  \caption{Single-MMKG transfer experiments following the setting proposed by IndMKG \cite{IndMKG}. A$\rightarrow$B denotes training on dataset A and testing on dataset B. The abbreviation for MMKG is as follows: YAGO15K(A), MKG-Y(B), MKG-W(C), WN18RR++(D), FB15K-237(E). We compare {\model} with two baselines under both zero-shot and fine-tuning settings.} 
  \label{fig::transfer}
  \vspace{-16pt}
\end{figure}
\par \textbf{(2). Comparison with IndMKG}. The above experiments mainly validate the effectiveness of {\model} in the standard KGFM setting. We now turn to a more challenging evaluation that explicitly stresses \emph{transferability} across different MMKGs, and compare {\model} with transferable MMKGR methods such as IndMKG~\cite{IndMKG}.
Following the protocol in the IndMKG paper, we conduct single-source transfer experiments between pairs of MMKGs, i.e., we pre-train on dataset A and directly perform KGR on dataset B without any additional training on B (A$\rightarrow$B). Due to the lack of fully functional released code, we keep their experimental configurations unchanged and directly reuse the experimental settings and reported results from IndMKG to ensure a fair and comparable evaluation.

\par Under this setting, {\model} achieves strong overall performance, outperforming IndMKG and KG-ICL in most zero-shot and fine-tuning comparisons (Figure~\ref{fig::transfer}). The gains are particularly pronounced in the zero-shot regime, where the model must rely purely on its pre-trained multi-modal and structural priors to adapt to a new graph. This highlights that {\model} is not merely competitive on in-domain benchmarks, but also exhibits strong cross-dataset and cross-graph transfer, which is crucial for practical deployment on diverse real-world KGs.

\subsection{Transductive MMKGR Experiments}
The main experiments primarily evaluate cross-graph transferability. To further assess the effectiveness of {\model} under the standard MMKGR setting, we additionally conduct transductive experiments where training and evaluation are performed on the same MMKG.
\begin{figure}[t]
  \centering
  \includegraphics[width=\linewidth]{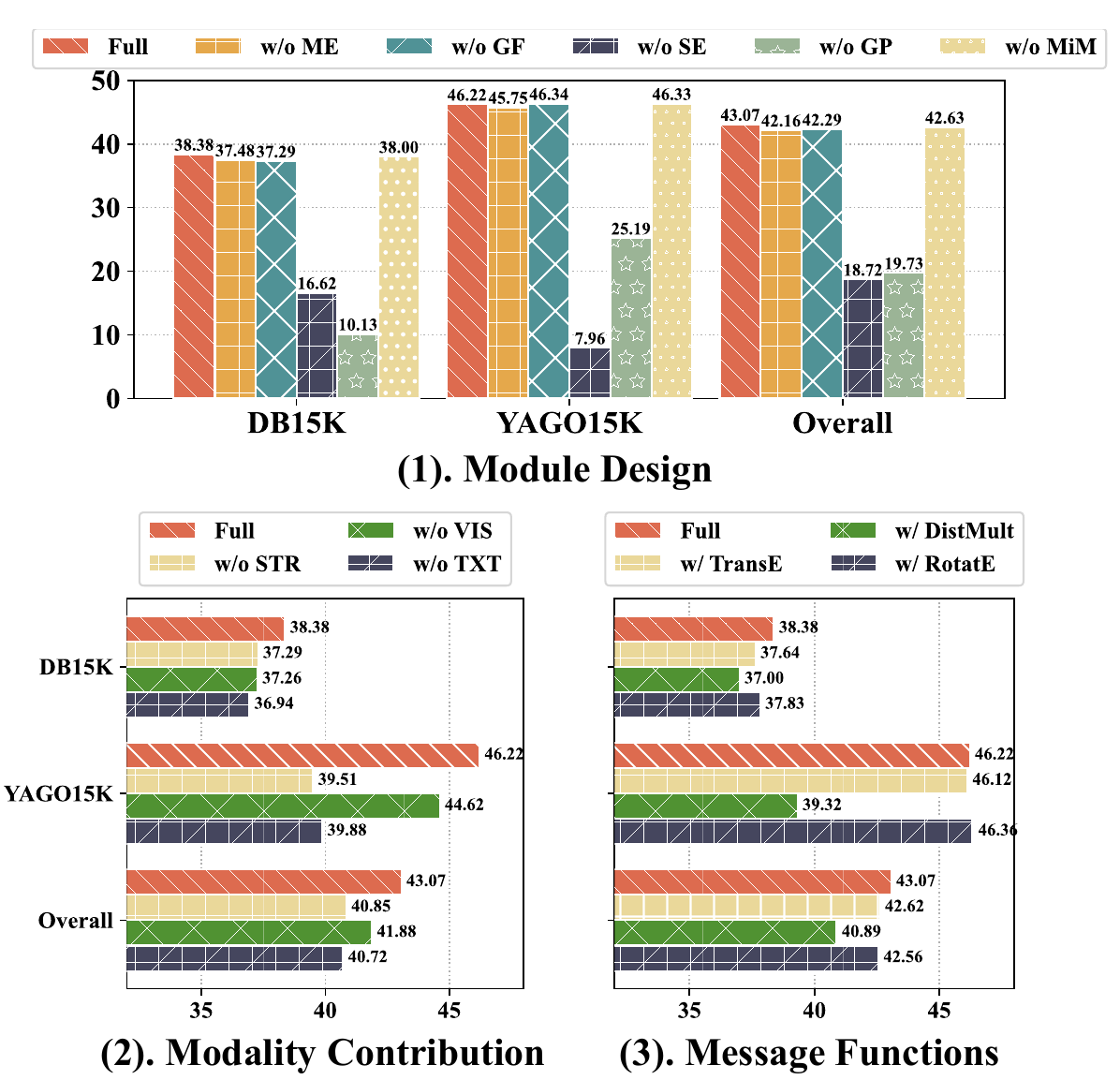}
  \caption{Ablation study on module design, modality contribution, and message function selections. The ablation study is conducted on 5 MMKGs separately, including DB15K, MKG-W, MKG-Y, YAGO15K, and WN18RR++. Overall results represent the average. We report the MRR results.} 
  \label{fig::ablation}
  \vspace{-16pt}
\end{figure}
\par We present the results on three benchmarks in Table \ref{table::transductive_mmkgr}. The results indicate that {\model} consistently outperforms traditional MMKGR methods in terms of overall performance, with especially notable gains on MKG-W and MKG-Y while maintaining a competitive performance on DB15K. On average, {\model} achieves the best results across most metrics. A more fine-grained inspection reveals that {\model} shows larger improvements on Hit@1, while being slightly weaker on Hit@10 compared to some baselines. Since all baselines adopt separate entity embeddings, whereas {\model} discards this design and relies purely on token-level multi-modal modeling, this suggests a trade-off: the lack of dedicated entity embeddings may moderately affect coarse-grained ranking, but does not hinder—and may even benefit—precise ranking. In summary, the transductive MMKGR experiments show that token-level modeling of all modal information yields positive gains for MMKGR.

\subsection{Ablation Study and Further Exploration}
\begin{figure*}[]
  \centering
  \includegraphics[width=\linewidth]{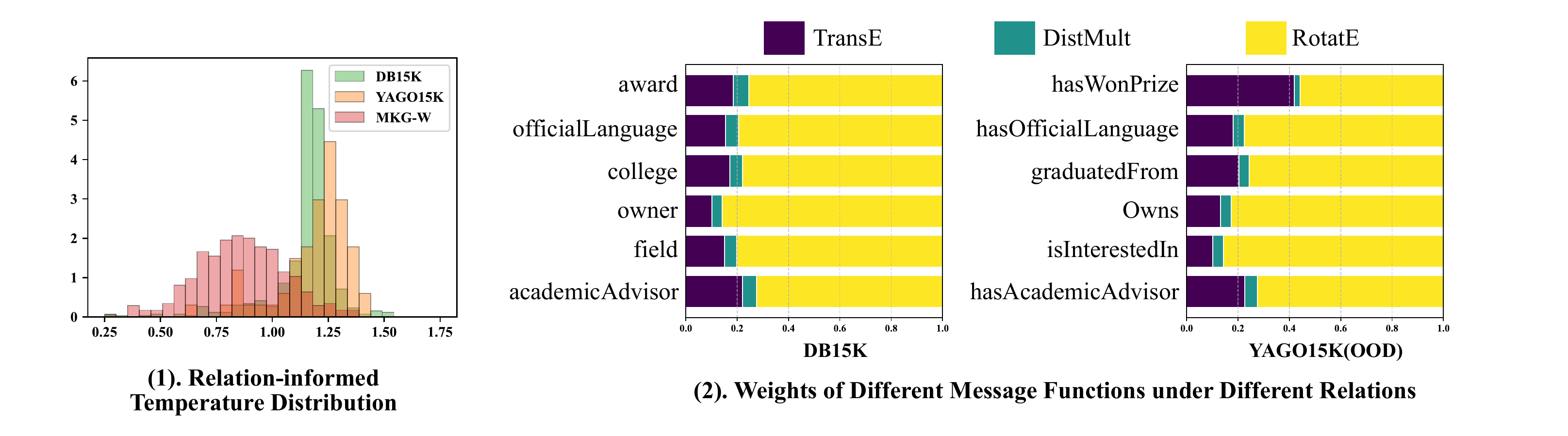}
  \vspace{-16pt}
  \caption{Case study on the mixture-of-message modules. We visualize the distribution of temperature $\gamma_i$ for different relations and the weights for the three message functions under different query relations. Note that YAGO15K is an OOD dataset.} 
  \label{fig::case_study}
  \vspace{-12pt}
\end{figure*}
We further conduct an ablation study to validate the reasonability of our design. The ablation study consists of three aspects: the overall module design, the modality contribution, and the message function selection.
\par As presented in Figure~\ref{fig::ablation}(1), each key module in {\model} positively contributes to the final performance. Although components such as GF do not yield consistent gains on every single dataset, their averaged effect over the 5 MMKGs is clearly beneficial. The results also indicate that SE and GP, the two GNN modules, still bring the largest improvements, underscoring that structural patterns and subgraph contexts are still significant in MMKGR, and multi-modal information plays complementary roles. Importantly, these multi-modal representations are foundational for {\model}: without them, structure-aware modules such as GP would lack informative node and relation features to exploit.
\par Figure~\ref{fig::ablation}(2) further examines the contribution of each modality by removing one modality at a time. All three modalities are useful, with textual information generally providing the strongest signal, while the relative importance of visual and textual cues varies across MMKGs. It is worth noting that w/o SE and w/o structural tokens represent two distinct experimental settings, yet both collectively confirm that structural information in the subgraph contexts is still the dominant modality for MMKGR tasks.

\par Next, we conducted further investigations into the message function within the GP module, exploring three classic functions: TransE, DistMult, and RotatE. We observe that the MiM module, which aggregates these heterogeneous messages via a MoE mechanism, achieves the best overall performance across datasets. Although a single message function may occasionally outperform MiM on a specific MMKG, MiM offers more stable and robust gains when averaged over multiple graphs, suggesting that different relation patterns are better captured by different message types.

\begin{table}[h]
\centering
\begin{tabular}{ccc}
\toprule
\textbf{Method} & \textbf{\#Params} & \textbf{Time Cost} \\
\midrule
MyGO & 45.63M &  57.76s \\
MoMoK & 72.40M & 33.38s \\
ULTRA & 0.169M & 31.12s  \\
KG-ICL & 0.054M & 49.13s  \\
MOTIF & 0.181M &  >500s \\
{\model} & 0.248M & 58.41s  \\
\bottomrule
\end{tabular}
\vspace{-8pt}
\caption{The efficiency analysis experiment on DB15K.}
\label{figure::efficiency}
\vspace{-12pt}
\end{table}
\subsection{Efficiency Analysis}
{\model} differs from traditional MMKGR models in that it does not rely on separate learnable entity and relation embedding tables. For traditional MMKGR methods trained on single MMKG, it has $O((|\mathcal{E}|+|\mathcal{R}|)\times d)$ learnable parameters where $d$ is the embedding dimension. Instead, it is composed of several neural network components. It is worth noting that the token embedding tables for text and image modalities are fixed during training. Consequently, the total number of trainable parameters in {\model} is substantially smaller than that of conventional MMKGR models that maintain large, fully learnable entity and relation embeddings. In Figure~\ref{figure::efficiency}, we present quantitative analysis on the learnable parameter amounts and inference time cost of different models. Compared to the traditional MMKGR methods, {\model} still has significantly fewer parameters, while the increase in parameters is very limited compared to other unimodal KGFMs. In terms of inference speed, there is no order-of-magnitude difference compared to other methods, which demonstrates the efficiency of {\model}.
\subsection{Case Study and Visualized Analysis}
To provide a more intuitive understanding of our design, we conduct a case study focusing on the GP module in {\model}. For the proposed mixture-of-messages mechanism, we visualize two key components: (i) the distribution of relation-guided temperature values, and (ii) the weights assigned to the three base message functions.
\par First, we visualized the temperature distribution in Figure~\ref{fig::case_study}(1). Following the KGFM configuration outlined earlier, we conducted pre-training on the DB15K/MKG-W/MKG-Y and FB15K-237(v1) datasets. For this temperature distribution visualization, we selected two in-distribution datasets and one out-of-distribution (OOD) dataset, YAGO15K. We visualized the mean logits before the sigmoid obtained for different relations across these datasets. The distributions differ noticeably across datasets, indicating that the relation-guided temperature adapts to the relation patterns of each graph. Moreover, the temperature distribution on the OOD YAGO15K dataset is clearly distinct from those on the pre-training datasets, yet still leads to performance gains, suggesting that this mechanism can generalize to unseen relational configurations.
\par We also visualized the weights of the three message functions in MiM. Several patterns emerge: first, the three weights for different relations show distinct differences, yet all exhibit relatively low dependence on distmult. The variations primarily concentrate on the TransE and RotatE message functions. Second, message weight distributions for similar relations across different datasets exhibit certain similarities. For example, the \textbf{\textit{academicAdvisor}} relation in DB15K and the \textbf{\textit{hasAcademicAdvisor}} relation in YAGO15K have very similar weight distributions. This emerges even though relation texts are not used during pre-training, indicating that {\model} captures transferable relation patterns and reuses them on OOD datasets. In other words, the message mixture learned during pre-training exhibits strong relational generalization on new MMKGs, which is crucial for robust MMKGR under distribution shifts.

\section{Conclusion}
In this paper, we focus on building a foundation model for MMKGR. We propose a new token-based framework {\model} to model the multi-modality (structure/visual/text) in MMKGs as discrete tokens. We first propose the token modeling process for three modalities and further propose a hierarchical architecture to accomplish MMKGR with the multi-modal fine-grained tokens with mixture-of-messages. Extensive experiments on 17 MMKGs under transductive, inductive, and fully-inductive settings demonstrate that {\model} consistently outperforms existing KGFM and MMKGR baselines. In-depth explorations are made to further show the rationality and mechanism interpretability of {\model}. In the future, the integration of more MMKG tasks like entity linking and entity alignment towards a unified multi-modal KGFM would be further explored.

\bibliography{aaai2027}
\clearpage
\appendix
\section{Experiments}

\subsection{Dataset Information}
\label{appendix::dataset}
As presented in Table \ref{table::datasets_full}, our MMKG datasets used in this paper can be divided into three categories:
\begin{itemize}
    \item \textbf{Transductive MMKG Datasets}: DB15K \cite{MMKG}, MKG-W \cite{MMRNS}, MKG-Y \cite{MMRNS}, YAGO15K \cite{MMKG}, WN9 \cite{xie_image-embodied_2017-IKRL}, WN18RR++ \cite{lee_vista_2023-VISTA}, FB15K-237-10\%/20\%/50\% \cite{FB15K-237-sparse}. These datasets have fixed entity and relation sets during training and evaluation.
    \item \textbf{Inductive MMKG Datasets}: Four inductive splits of FB15K-237(v1/v2/v3/v4) from GraIL \cite{GRAIL}. These datasets consist of new entities in the evaluation stage.
    \item \textbf{Fully-inductive MMKG Datasets}: Four fully-inductive splits of FB15K-237(-25/-50/-75/-100) from InGram \cite{INGRAM}, which consists of both new entities and relations in the evaluation stage, which would be more challenging for the model's generalization capability.
\end{itemize}
The distinguishing criterion among the three types of datasets is whether their train/valid/test sets contain overlapping entities and relations. For transductive datasets, the entities and relations are consistent across the three splits. For inductive datasets, new entities are introduced in the test sets. For fully-inductive datasets, both new entities and relations would be introduced. Therefore, traditional KGR methods that learn separate embeddings for every entity and relation would not work for the inductive and fully-inductive datasets.

\par In the transductive experiments, we employ DB15K/MKG-W/MKG-Y as the benchmark datasets to present the transductive capability of the models. We then conduct KGFM experiments on all the mentioned datasets. We pre-train the models on DB15K/MKG-W/MKG-Y/FB15K-237-v1 and evaluate the model performance on all datasets with both zero-shot and fine-tuning settings. Given the limited availability of inductive and fully-inductive MMKG datasets at present, we can only utilize the FB15K-237 dataset, which readily provides multi-modal information. \textbf{It is important to note that we can not directly use the complete FB15K-237 dataset as pre-training data, as this would cause the model to directly leak data from the inductive and fully-inductive experiments.} Existing KGFM work has also followed this approach and employed FB15K-237(v1) as pre-training data.
\par Another noteworthy point is that different MMKGs are not necessarily completely isolated. They may share some overlapping entities or triples. However, such overlap is unavoidable within KGFM, which does not mean data leakage. The corresponding multi-modal information may also exhibit minor similarities. Overall, we conduct pre-training and fine-tuning across different MMKGs to demonstrate KGFM's ability to jointly master and apply transferable features across structural, visual, and textual domains.
\subsection{Baseline Methods}
\label{appendix::baseline}

For transdutive MMKGC experiments, the baselines we used for performance comparation includes IKLR \cite{xie_image-embodied_2017-IKRL}, TBKGC \cite{sergieh_multimodal_2018-TBKGC}, TransAE \cite{wang_multimodal_2019-TransAE}, MMKRL \cite{DBLP:journals/apin/LuWJHL22-MMKRL}, RSME \cite{wang_is_2021-RSME}, VBKGC \cite{DBLP:journals/corr/abs-2209-07084-VBKGC}, OTKGE \cite{cao_otkge_2022-OTKGE}, MMRNS \cite{MMRNS}, IMF \cite{li_imf_2023-IMF}, MANS \cite{DBLP:conf/ijcnn/ZhangCZ23-MANS}, QEB \cite{DBLP:conf/mm/WangMCML023-TIVA}, VISTA \cite{lee_vista_2023-VISTA}, NATIVE \cite{NATIVE}, AdaMF \cite{AdaMF}, SNAG \cite{SNAG}, MyGO \cite{MyGO}, MoMoK \cite{MoMoK}, K-ON \cite{K-ON}, MCKGC \cite{MCKGC}, APKGC \cite{APKGC}, LMBKGC \cite{LMBKGC}.
\par These baseline methods are all classic MMKGR baselines in recent years. Their common feature is that they are designed for transductive MMKGR tasks, where both training and inference are performed on the same fixed entity and relation set to make predictions for different triples. Therefore, nearly all approaches design separate entity/relation embedding tables, learning distinct structural embeddings for each entity and relation, supplemented by multi-modal information fusion to enhance the model. To achieve multimodal fusion, they designed numerous innovative architectures, including attention, transformer, MoE, diffusion, and more.

\subsection{Hyper-parameter Settings}
\label{appendix::parameters}
We implement the experiments with PyTorch \cite{pytorch} and PyTorch-Geometric \cite{pyg}. In the main experiments, we set the training epochs to 30 with a batch size of 256. The hidden dimensions of SE and GP are set to 32 for all experiments. The learning rate is set to 0.0005 using the Adam \cite{DBLP:journals/corr/KingmaB14-Adam} optimizer.
\par For token process, we employ BERT \cite{BERT} and BEiT \cite{BEIT} as text/image tokenizers respectively. BERT tokenizer is the byte pair encoding strategy to tokenize texts into subwords. BEiT employs a VQ-VAE \cite{VQ-VAE} to process one image into 14$\times$14 discrete visual tokens. The embedding dimensions of the textual and visual tokens are 768 and 32, respectively. In the main experiments, we set $N_{vis}, N_{txt}$ to 8. The GNN layers in SE and GP consist of 6 layers.

\subsection{Full KGFM Experiment Results}
The full KGFM experiment results are presented in Table~\ref{table::kgfm_full}, in the original version, we categorized FB15K-237(v1) as a transductive dataset when computing the aggregated results because it was included in the pre-training data. Following the convention adopted by most prior studies, we classify all four FB15K-237 splits (v1–v4) as inductive datasets. Consequently, the Transductive and Inductive averages in Table 2 have been updated using 9 and 4 datasets respectively. This recategorization only changes the two category-level averages; the results on each individual dataset and the overall performance across all 17 datasets remain unchanged.
\subsection{More MMKGR Results Compared with IndMKG}
\label{appendix::indmkg_results}
We conduct more experiments with IndMKG \cite{IndMKG}, an inductive MMKGR method that follows the pre-training and fine-tuning paradigm. However, the experimental setup they employed involved transfer learning between individual MMKGs—specifically, pre-training on one MMKG and then fine-tuning on another for prediction. Strictly speaking, this approach is not entirely inductive, as the pre-trained MMKG and the fine-tuned MMKG may share overlapping entities. Nevertheless, such experiments effectively demonstrate the transfer capabilities of different models across different MMKGs.

\par The full experimental results are presented in Table~\ref{table::indmkg_full}. Here, we provide more inductive KGR and MMKGR baselines like InGram \cite{INGRAM}, IMF \cite{li_imf_2023-IMF}, MoCi \cite{MoCi}, ULTRA \cite{ULTRA}, HyRel \cite{HyRel}, IndMKG \cite{IndMKG}, and KG-ICL \cite{KGICL} for comparison.

\subsection{Ablation Study}
\begin{figure}[]
  \centering
  \includegraphics[width=\linewidth]{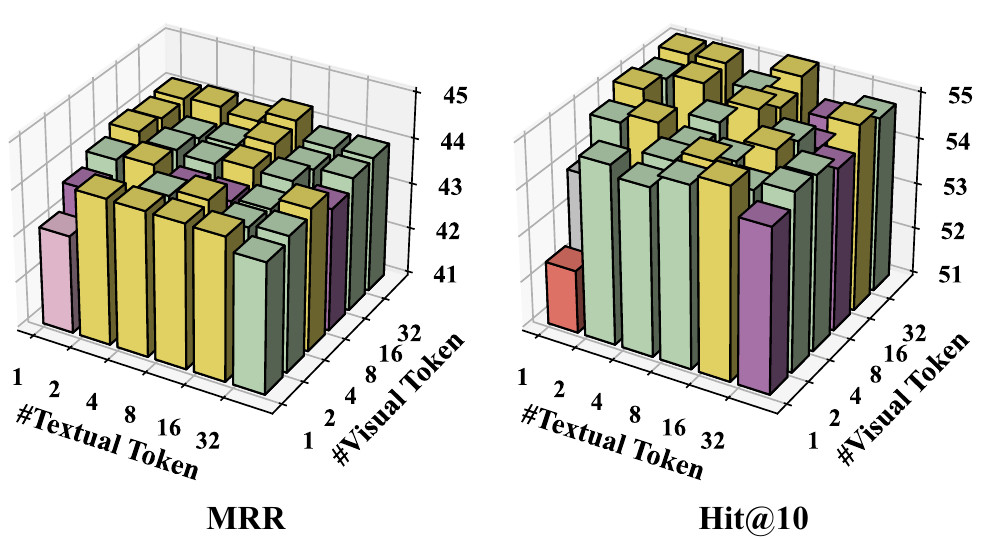}
  \caption{The influence of text/visual token number.} 
  \label{fig::token_num}
\end{figure}
\par We study the effect of the number of visual and textual tokens $N_{vis}, N_{txt}$ in Figure~\ref{fig::token_num}. A fundamental principle is that model performance tends to increase gradually as the token count grows. However, considering the balance between efficiency and performance, we typically opt for a relatively small value such as 8 or 16. Note that the structural modality is represented by a single token whose features are directly aggregated by SE, leading to a much higher information density per structural token than for visual or textual tokens.

\subsection{Efficiency and Parameter Analysis of {\model}}
\label{appendix::parameter_efficiency}
As summarized in the main paper, {\model} differs from traditional MMKGR models in that it does not rely on separate learnable entity and relation embedding tables. For traditional MMKGR methods trained on single MMKG, it has $O((|\mathcal{E}|+|\mathcal{R}|)\times d)$ learnable parameters where $d$ is the embedding dimension. Instead, it is composed of a set of neural network components such as GNN layers, Transformer blocks, and projection layers, and learns a transferable model by training the parameters within these modules. It is worth noting that the token embedding tables for text and image modalities are kept fixed during training and are not updated via gradient descent. Consequently, the total number of trainable parameters in {\model} is substantially smaller than that of conventional MMKGR models that maintain large, fully learnable entity and relation embeddings. In addition, the number of parameters in {\model} is fixed and does not grow with the size or the number of MMKG datasets. The same set of parameters is used for training and inference across multiple datasets, which is consistent with the design philosophy of existing KGFMs. The key difference is that, on top of this foundation-model style parameter sharing, {\model} further incorporates dedicated designs for multi-modal fusion. Using the DB15K dataset as an example, the number of trainable parameters for several key components involved in our method are as follows:
\begin{itemize}
    \item Traditional MMKGR methods: MoMoK 72.4M, MyGO 45.63M
    \item KGFM methods: ULTRA 0.169M, KG-ICL 0.054M, MOTIF 0.181M
    \item {\model} (ours): 0.248M
\end{itemize}
We observe that, compared with existing KGFMs, {\model} indeed introduces a larger number of parameters. However, its parameter scale is still below 1\% of that of traditional MMKGR methods. Considering that {\model} additionally incorporates multi-modal information, the extra trainable parameters introduced for multi-modal fusion remain relatively small. Overall, {\model} is parameter-efficient while achieving the benefits of multi-modality.


\begin{table*}[t]
\caption{The full results of the KGFM experiments on 17 MMKG datasets compared with supervised SOTA results and other KGFM methods. The supervised SOTA results are mainly provided by KG-ICL \cite{KGICL}.}
\label{table::kgfm_full}
\centering
\resizebox{0.85\textwidth}{!}{
\begin{tabular}{cc|c|c|cc|cc|cc|cc}
\toprule
\multicolumn{2}{c|}{\multirow{2}{*}{\textbf{Dataset}}} & \multirow{2}{*}{\textbf{Metrics}} & \multirow{2}{*}{\textbf{\begin{tabular}[c]{@{}c@{}}Supervised \\      SOTA\end{tabular}}} & \multicolumn{2}{c|}{\textbf{ULTRA}} & \multicolumn{2}{c|}{\textbf{KG-ICL}} & \multicolumn{2}{c|}{\textbf{MOTIF}} & \multicolumn{2}{c}{\textbf{{\model}}} \\
\multicolumn{2}{c|}{} &  &  & \textbf{pre-train} & \textbf{fine-tune} & \textbf{pre-train} & \textbf{fine-tune} & \textbf{pre-train} & \textbf{fine-tune} & \textbf{pre-train} & \textbf{fine-tune} \\
\midrule
\multirow{19}{*}{\rotatebox{90}{\textbf{Transductive}}} & \multirow{2}{*}{\textbf{DB15K}} & \textbf{MRR} & 37.23 & 32.98 & \textbf{40.72} & 32.19 & 31.95 & 29.48 & 38.59 & 38.38 & \underline{38.62} \\
 &  & \textbf{Hit@10} & 54.71 & 48.36 & \textbf{56.59} & 45.11 & 45.23 & 46.34 & 55.76 & 52.73 & \underline{53.02} \\
 & \multirow{2}{*}{\textbf{MKG-W}} & \textbf{MRR} & 38.46 & 34.41 & 37.77 & 34.10 & 34.07 & 35.82 & 38.07 & \underline{39.80} & \textbf{40.55} \\
 &  & \textbf{Hit@10} & \underline{51.46} & 45.68 & 49.13 & 39.23 & 43.37 & 46.99 & 49.60 & 50.71 & \textbf{51.70} \\
 & \multirow{2}{*}{\textbf{MKG-Y}} & \textbf{MRR} & 40.03 & 35.62 & 39.35 & 40.76 & 40.61 & 33.94 & 39.44 & \textbf{43.72} & \underline{43.65} \\
 &  & \textbf{Hit@10} & \textbf{50.81} & 42.09 & 45.86 & 46.94 & 47.03 & 44.04 & 46.00 & 49.77 & \underline{50.06} \\
 & \multirow{2}{*}{\textbf{YAGO15K}} & \textbf{MRR} & 43.03 & 41.06 & 45.29 & 43.50 & 44.48 & 39.37 & 44.74 & \underline{46.22} & \textbf{50.31} \\
 &  & \textbf{Hit@10} & 54.49 & 54.20 & 58.16 & 53.43 & 54.69 & 54.21 & 58.14 & \underline{59.99} & \textbf{62.89} \\
 & \multirow{2}{*}{\textbf{WN9}} & \textbf{MRR} & 92.30 & 26.72 & 90.23 & 94.68 & \underline{95.15} & 34.56 & 91.20 & 94.31 & \textbf{95.15} \\
 &  & \textbf{Hit@10} & 94.70 & 50.26 & 91.96 & 95.15 & \underline{95.15} & 63.84 & 92.57 & 95.00 & \textbf{95.15} \\
 & \multirow{2}{*}{\textbf{WN18RR++}} & \textbf{MRR} & \underline{55.26} & 38.36 & 53.65 & 41.25 & 43.37 & 49.18 & 54.16 & 47.22 & \textbf{57.99} \\
 &  & \textbf{Hit@10} & \underline{67.55} & 54.17 & 64.15 & 48.28 & 51.53 & 58.64 & 64.26 & 56.03 & \textbf{67.87} \\
 & \multirow{2}{*}{\textbf{FB15K-237-10\%}} & \textbf{MRR} & 21.90 & 24.80 & 25.40 & \textbf{27.40} & 26.00 & 23.60 & 25.40 & 26.10 & \underline{26.18} \\
 &  & \textbf{Hit@10} & 33.70 & 39.80 & 41.10 & \textbf{43.30} & 41.60 & 38.40 & 41.10 & 41.52 & \underline{41.70} \\
 & \multirow{2}{*}{\textbf{FB15K-237-20\%}} & \textbf{MRR} & 24.70 & 27.20 & 27.40 & \underline{28.50} & 28.40 & 25.90 & 27.30 & 27.89 & \textbf{28.65} \\
 &  & \textbf{Hit@10} & 39.10 & 43.60 & 44.50 & 45.40 & \underline{45.60} & 42.20 & 44.40 & 44.45 & \textbf{45.95} \\
 & \multirow{2}{*}{\textbf{FB15K-237-50\%}} & \textbf{MRR} & 29.30 & 32.40 & 32.50 & \underline{32.90} & 32.40 & 31.20 & 32.30 & 31.11 & \textbf{33.99} \\
 &  & \textbf{Hit@10} & 45.80 & \underline{52.60} & 52.80 & 52.00 & 49.90 & 50.80 & 52.30 & 50.43 & \textbf{52.97} \\
\midrule
\multirow{8}{*}{\rotatebox{90}{\textbf{Inductive}}} & \multirow{2}{*}{\textbf{FB15K-237(V1)}} & \textbf{MRR} & 45.70 & 49.80 & 50.90 & 52.00 & \underline{53.10} & 50.30 & 53.00 & 51.71 & \textbf{53.63} \\
 &  & \textbf{Hit@10} & 58.90 & 65.60 & 67.00 & 67.80 & 70.00 & 69.20 & \textbf{70.20} & 68.05 & \underline{70.00} \\
 & \multirow{2}{*}{\textbf{FB15K-237(V2)}} & \textbf{MRR} & 51.00 & 51.20 & 52.40 & 56.50 & \underline{56.80} & 51.10 & 55.70 & 55.45 & \textbf{57.21} \\
 &  & \textbf{Hit@10} & 67.20 & 70.00 & 71.00 & 74.90 & \textbf{76.80} & 71.60 & 74.40 & 74.69 & \underline{76.05} \\
 & \multirow{2}{*}{\textbf{FB15K-237(V3)}} & \textbf{MRR} & 47.60 & 49.10 & 50.40 & 53.50 & \underline{53.70} & 50.00 & 51.90 & 53.34 & \textbf{54.27} \\
 &  & \textbf{Hit@10} & 63.70 & 65.40 & 66.30 & 69.50 & \textbf{70.40} & 69.20 & 68.40 & 69.71 & \underline{70.12} \\
 & \multirow{2}{*}{\textbf{FB15K-237(V4)}} & \textbf{MRR} & 46.60 & 48.60 & 49.60 & 51.30 & \underline{52.50} & 48.70 & 50.80 & 51.75 & \underline{52.84} \\
 &  & \textbf{Hit@10} & 64.50 & 67.70 & 68.40 & 69.90 & 70.60 & 67.70 & 69.50 & \underline{70.75} & \textbf{71.17} \\
\midrule
\multirow{8}{*}{\rotatebox{90}{\textbf{Fully-Inductive}}} & \multirow{2}{*}{\textbf{FB-25}} & \textbf{MRR} & 22.30 & 38.80 & 38.30 & 39.60 & \textbf{43.40} & 38.40 & 37.60 & \underline{42.81} & 41.91 \\
 &  & \textbf{Hit@10} & 37.10 & 64.00 & 63.50 & 65.60 & \underline{69.40} & 64.00 & 62.10 & \textbf{69.51} & 69.00 \\
 & \multirow{2}{*}{\textbf{FB-50}} & \textbf{MRR} & 18.90 & 33.80 & 33.40 & 34.10 & \textbf{38.40} & 33.80 & 33.40 & \underline{38.14} & 37.76 \\
 &  & \textbf{Hit@10} & 32.50 & 54.30 & 53.80 & 55.90 & 59.80 & 54.60 & 53.20 & \textbf{60.36} & \underline{59.90} \\
 & \multirow{2}{*}{\textbf{FB-75}} & \textbf{MRR} & 11.70 & 40.30 & 40.00 & 41.80 & \textbf{45.80} & 39.90 & 38.90 & 44.00 & \underline{44.08} \\
 &  & \textbf{Hit@10} & 21.80 & 60.40 & 59.80 & 63.30 & \textbf{66.40} & 61.40 & 59.20 & \underline{65.12} & 64.44 \\
 & \multirow{2}{*}{\textbf{FB-100}} & \textbf{MRR} & 13.30 & 44.90 & 44.40 & 48.70 & \textbf{49.90} & 42.80 & 41.80 & 48.80 & \underline{49.12} \\
 &  & \textbf{Hit@10} & 27.10 & 64.20 & 64.30 & 69.40 & \textbf{70.30} & 62.80 & 61.40 & \underline{69.60} & 69.36 \\
\bottomrule
\end{tabular}
}
\end{table*}

\begin{table*}[]
\centering
\caption{Single MMKG transfer experiments.}
\label{table::indmkg_full}
\centering
\resizebox{\textwidth}{!}{
\begin{tabular}{c|ccc|ccc|ccc|ccc|ccc}
\toprule
\multicolumn{16}{c}{\textbf{\textit{Source KG: YAGO15K}}} \\
\midrule
\multirow{2}{*}{\textbf{Method}} & \multicolumn{3}{c|}{\textbf{MKG-Y}} & \multicolumn{3}{c|}{\textbf{MKG-W}} & \multicolumn{3}{c|}{\textbf{WN18RR++}} & \multicolumn{3}{c|}{\textbf{FB15K-237}} & \multicolumn{3}{c}{\textbf{Average}} \\
 & \textbf{MRR} & \textbf{Hit@1} & \textbf{Hit@10} & \textbf{MRR} & \textbf{Hit@1} & \textbf{Hit@10} & \textbf{MRR} & \textbf{Hit@1} & \textbf{Hit@10} & \textbf{MRR} & \textbf{Hit@1} & \textbf{Hit@10} & \textbf{MRR} & \textbf{Hit@1} & \textbf{Hit@10} \\
\midrule
InGram & 8.30 & 5.10 & 12.10 & 0.60 & 0.20 & 1.30 & 3.10 & 1.90 & 4.30 & 1.30 & 0.40 & 2.60 & 3.33 & 1.90 & 5.08 \\
IMF & 0.46 & 0.21 & 0.71 & 9.59 & 8.87 & 10.35 & 2.15 & 1.88 & 2.58 & 2.10 & 1.65 & 2.97 & 3.58 & 3.15 & 4.15 \\
MoCi & 2.06 & 1.51 & 2.63 & 14.27 & 10.23 & 20.30 & 2.55 & 2.01 & 3.78 & 2.43 & 1.79 & 3.67 & 5.33 & 3.89 & 7.60 \\
ULTRA & 35.13 & 31.13 & 42.99 & 29.91 & 23.19 & 43.10 & 27.53 & 19.19 & 42.94 & 15.96 & 9.16 & 29.83 & 27.13 & 20.67 & 39.72 \\
HyRel & 10.40 & 6.30 & 13.40 & 2.30 & 0.50 & 4.10 & 4.30 & 2.10 & 8.40 & 6.10 & 2.30 & 9.30 & 5.78 & 2.80 & 8.80 \\
IndMKG(zero-shot) & 36.04 & 33.00 & 41.50 & 33.38 & 27.63 & 44.72 & 31.94 & 23.68 & 47.08 & 28.70 & 19.79 & 46.89 & 32.52 & 26.03 & 45.05 \\
IndMKG(fine-tune) & 39.58 & 36.20 & 45.83 & 37.15 & 31.30 & 48.80 & 53.51 & 49.90 & 60.70 & 35.14 & 27.10 & 51.80 & 41.35 & 36.13 & 51.78 \\
KG-ICL(zero-shot) & 39.10 & 35.30 & 46.58 & 31.37 & 26.06 & 41.37 & 31.01 & 23.21 & 45.72 & 20.58 & 13.93 & 33.63 & 30.52 & 24.63 & 41.83 \\
KG-ICL(fine-tune) & 40.26 & 36.84 & 46.60 & 33.72 & 28.63 & 43.26 & 41.34 & 37.83 & 47.99 & 32.36 & 24.77 & 47.52 & 36.92 & 32.02 & 46.34 \\
{\model}(zero-shot) & 42.60 & 38.94 & 49.00 & 36.93 & 30.88 & 48.21 & 29.67 & 20.74 & 46.89 & 33.86 & 25.44 & 50.35 & 35.77 & 29.00 & 48.61 \\
{\model}(fine-tune) & 43.83 & 40.50 & 49.89 & 40.48 & 34.55 & 51.50 & 57.46 & 52.74 & 67.17 & 39.19 & 30.82 & 55.56 & 45.24 & 39.65 & 56.03 \\
\midrule
\multicolumn{16}{c}{\textbf{\textit{Source KG: \textbf{WN18RR++}}}} \\
\midrule
\multirow{2}{*}{\textbf{Method}} & \multicolumn{3}{c|}{\textbf{MKG-W}} & \multicolumn{3}{c|}{\textbf{YAGO15K}} & \multicolumn{3}{c|}{\textbf{MKG-Y}} & \multicolumn{3}{c|}{\textbf{FB15K-237}} & \multicolumn{3}{c}{\textbf{Average}} \\

 & \textbf{MRR} & \textbf{Hit@1} & \textbf{Hit@10} & \textbf{MRR} & \textbf{Hit@1} & \textbf{Hit@10} & \textbf{MRR} & \textbf{Hit@1} & \textbf{Hit@10} & \textbf{MRR} & \textbf{Hit@1} & \textbf{Hit@10} & \textbf{MRR} & \textbf{Hit@1} & \textbf{Hit@10} \\
\midrule
InGram & 6.50 & 2.40 & 9.60 & 5.70 & 3.70 & 9.10 & 10.30 & 8.60 & 13.50 & 7.50 & 3.50 & 11.30 & 7.50 & 4.55 & 10.88 \\
IMF & 8.75 & 8.17 & 9.41 & 6.50 & 5.22 & 8.78 & 0.50 & 0.26 & 0.81 & 2.13 & 1.66 & 3.00 & 4.47 & 3.83 & 5.50 \\
MoCi & 13.56 & 10.01 & 19.85 & 9.33 & 4.78 & 17.84 & 2.14 & 1.63 & 2.66 & 2.28 & 1.65 & 3.59 & 6.83 & 4.52 & 10.99 \\
ULTRA & 17.68 & 13.19 & 26.12 & 17.76 & 13.52 & 24.87 & 13.52 & 9.29 & 21.19 & 10.89 & 5.60 & 21.07 & 14.96 & 10.40 & 23.31 \\
HyRel & 7.90 & 3.50 & 11.30 & 8.90 & 3.90 & 11.10 & 10.60 & 8.90 & 14.00 & 9.10 & 6.70 & 14.40 & 9.13 & 5.75 & 12.70 \\
IndMKG(zero-shot) & 23.63 & 18.21 & 33.70 & 18.04 & 11.22 & 31.70 & 18.04 & 12.67 & 30.05 & 13.46 & 7.31 & 25.78 & 18.29 & 12.35 & 30.31 \\
IndMKG(fine-tune) & 37.29 & 31.40 & 48.60 & 44.13 & 39.00 & 54.10 & 39.46 & 35.70 & 46.00 & 36.87 & 27.90 & 55.10 & 39.44 & 33.50 & 50.95 \\
KG-ICL(zero-shot) & 33.50 & 29.33 & 41.70 & 31.38 & 21.58 & 46.51 & 34.14 & 29.85 & 42.66 & 11.62 & 8.09 & 18.17 & 27.66 & 22.21 & 37.26 \\
KG-ICL(fine-tune) & 32.67 & 27.36 & 42.63 & 43.44 & 37.83 & 53.44 & 38.76 & 34.55 & 46.56 & 31.88 & 24.06 & 47.44 & 36.69 & 30.95 & 47.52 \\
{\model}(zero-shot) & 27.49 & 21.82 & 39.39 & 34.50 & 30.14 & 42.55 & 35.15 & 30.19 & 44.89 & 12.13 & 7.29 & 22.06 & 27.32 & 22.36 & 37.22 \\
{\model}(fine-tune) & 40.59 & 34.58 & 51.57 & 49.87 & 42.91 & 62.54 & 43.63 & 40.37 & 49.81 & 37.20 & 28.81 & 53.46 & 42.82 & 36.67 & 54.35 \\
\midrule
\multicolumn{16}{c}{\textbf{\textit{Source KG: \textbf{MKG-Y}}}} \\
\midrule
\multirow{2}{*}{\textbf{Method}} & \multicolumn{3}{c|}{\textbf{MKG-W}} & \multicolumn{3}{c|}{\textbf{YAGO15K}} & \multicolumn{3}{c|}{\textbf{WN18RR++}} & \multicolumn{3}{c|}{\textbf{FB15K-237}} & \multicolumn{3}{c}{\textbf{Average}} \\
 & \textbf{MRR} & \textbf{Hit@1} & \textbf{Hit@10} & \textbf{MRR} & \textbf{Hit@1} & \textbf{Hit@10} & \textbf{MRR} & \textbf{Hit@1} & \textbf{Hit@10} & \textbf{MRR} & \textbf{Hit@1} & \textbf{Hit@10} & \textbf{MRR} & \textbf{Hit@1} & \textbf{Hit@10} \\
 \midrule
InGram & 6.40 & 3.10 & 10.50 & 1.20 & 0.30 & 1.80 & 0.40 & 0.10 & 1.30 & 6.40 & 2.30 & 9.50 & 3.60 & 1.45 & 5.78 \\
IMF & 9.61 & 8.89 & 10.54 & 4.91 & 2.43 & 7.37 & 2.15 & 1.88 & 2.58 & 2.08 & 1.62 & 2.88 & 4.69 & 3.71 & 5.84 \\
MoCi & 13.77 & 10.21 & 19.75 & 9.08 & 4.87 & 17.96 & 2.45 & 1.97 & 3.81 & 2.54 & 1.79 & 4.12 & 6.96 & 4.71 & 11.41 \\
ULTRA & 24.46 & 17.53 & 38.51 & 34.13 & 28.75 & 43.03 & 27.01 & 17.86 & 42.15 & 11.58 & 7.19 & 20.35 & 24.30 & 17.83 & 36.01 \\
HyRel & 8.70 & 4.70 & 10.90 & 6.50 & 1.40 & 8.40 & 5.10 & 1.60 & 7.70 & 10.70 & 5.10 & 13.50 & 7.75 & 3.20 & 10.13 \\
IndMKG(zero-shot) & 30.33 & 23.81 & 42.54 & 36.88 & 30.70 & 48.84 & 31.33 & 22.91 & 46.95 & 17.72 & 10.75 & 32.94 & 29.07 & 22.04 & 42.82 \\
IndMKG(fine-tune) & 36.79 & 30.40 & 49.10 & 44.72 & 38.60 & 55.90 & 53.73 & 49.70 & 62.30 & 36.80 & 27.10 & 56.20 & 43.01 & 36.45 & 55.88 \\
KG-ICL(zero-shot) & 18.16 & 11.43 & 32.84 & 34.66 & 28.03 & 46.74 & 27.79 & 17.95 & 45.23 & 9.25 & 5.80 & 15.60 & 22.47 & 15.80 & 35.10 \\
KG-ICL(fine-tune) & 32.61 & 27.29 & 42.95 & 43.19 & 37.64 & 53.10 & 28.18 & 16.59 & 47.16 & 33.09 & 25.07 & 49.01 & 34.27 & 26.65 & 48.06 \\
{\model}(zero-shot) & 30.96 & 24.94 & 42.17 & 39.08 & 32.45 & 48.15 & 32.59 & 23.52 & 48.15 & 14.97 & 8.76 & 27.13 & 29.40 & 22.42 & 41.40 \\
{\model}(fine-tune) & 40.00 & 33.19 & 51.42 & 49.81 & 42.71 & 62.83 & 57.99 & 53.11 & 67.47 & 37.30 & 29.21 & 53.23 & 46.28 & 39.56 & 58.74 \\
\bottomrule
\end{tabular}
}
\end{table*}

\end{document}